\documentclass[a4paper]{article}

\usepackage{INTERSPEECH2018}

\usepackage{graphicx}
\usepackage{amssymb,amsmath,bm}
\usepackage{textcomp}
\usepackage[inline]{enumitem}
\usepackage{graphics}
\usepackage{multirow}
\usepackage{hyperref}
\usepackage{xcolor}

\def\vec#1{\ensuremath{\bm{{#1}}}}

\def\bmu{{\bm \mu}}

\title{Unsupervised Adaptation with Interpretable Disentangled Representations \\for Distant Conversational Speech Recognition}

\name{Wei-Ning Hsu, Hao Tang, James Glass}
\address{Computer Science and Artificial Intelligence Laboratory \\
         Massachusetts Institute of Technology\\
         Cambridge, MA 02139, USA}
\email{\{wnhsu,haotang,glass\}@mit.edu}

\begin{document}

  \maketitle
  
  \begin{abstract}
  The current trend in automatic speech recognition is to leverage large amounts of labeled data to train supervised neural network models.
  Unfortunately, obtaining data for a wide range of domains to train robust models can be costly.
  However, it is relatively inexpensive to collect large amounts of unlabeled data from domains that we want the models to generalize to.
  In this paper, we propose a novel unsupervised adaptation method that learns to synthesize labeled data for the target domain from unlabeled in-domain data and labeled out-of-domain data.
  We first learn without supervision an interpretable latent representation of speech that encodes linguistic and nuisance factors (e.g., speaker and channel) using different latent variables.
  To transform a labeled out-of-domain utterance without altering its transcript, we transform the latent nuisance variables while maintaining the linguistic variables.
  To demonstrate our approach, we focus on a channel mismatch setting, where the domain of interest is distant conversational speech, and labels are only available for close-talking speech.
  Our proposed method is evaluated on the AMI dataset, outperforming all baselines and bridging the gap between unadapted and in-domain models by over 77\% without using any parallel data.
  \end{abstract}
  \noindent{\bf Index Terms}: unsupervised adaptation, distant speech recognition, unsupervised data augmentation, variational autoencoder

  % SECTION 1
  \section{Introduction}
  Distant speech recognition has greatly improved due to the recent advance in neural network-based acoustic models, which facilitates integration of automatic speech recognition (ASR) systems into hands-free human-machine interaction scenarios~\cite{li2017acoustic}.
  To build a robust acoustic model, previous work primarily focused on collecting labeled in-domain data for fully supervised training~\cite{zhang2016highway,hsu2016prioritized,kim2017residual}.
  However, in practice, it is expensive and laborious to collect labeled data for all possible testing conditions.
  In contrast, collecting large amount of unlabeled in-domain data and labeled out-of-domain data can be fast and economical.
  Hence, an important question arises for this scenario: \textit{how can we do unsupervised adaptation for acoustic models by utilizing labeled out-of-domain data and unlabeled in-domain data, in order to achieve good performance on in-domain data?}
  
  Research on unsupervised adaptation for acoustic models can be roughly divided into three categories:
  \begin{enumerate*}[label=(\arabic*)]
    \item constrained model adaptation~\cite{gales1998maximum, swietojanski2014learning, swietojanski2016learning},
    \item domain-invariant feature extraction~\cite{sun2017unsupervised, meng2017unsupervised, hsu2018extracting}, and
    \item labeled in-domain data augmentation by synthesis~\cite{Jaitly_vocaltract13, ko2015audio, hsu2017unsuperviseddomain}.
  \end{enumerate*}
  Among these approaches, data augmentation-based adaptation is favorable, because it does not require extra hyperparameter tuning for acoustic model training, and can utilize full model capacity by training a model with as much and as diverse a dataset as possible. 
  Another benefit of this approach is that data in their original domain are more intuitive to humans. In other words, it is easier for us to inspect and manipulate the data.
  Furthermore, with the recent progress on domain translation~\cite{hsu2017unsuperviseddomain, zhu2017unpaired, hsu2017unsupervised}, conditional synthesis of in-domain data without parallel data has become achievable, which makes data augmentation-based adaptation a more promising direction to investigate.

  Variational autoencoder-based data augmentation (VAE-DA) is a domain adaptation method proposed in~\cite{hsu2017unsuperviseddomain}, which pools in-domain and out-domain to train a VAE that learns factorized latent representations of speech segments. 
  To disentangle linguistic factors from nuisance ones in the latent space, statistics of the latent representations for each utterance are computed.
  By altering the latent representations of the segments from a labeled out-of-domain utterance properly according to the computed statistics, one can synthesize an in-domain utterance without changing the linguistic content using the trained VAE decoder.
  This approach shows promising results on synthesizing noisy read speech from clean speech. 
  However, it is non-trivial to apply this approach to conversational speech, because utterances tend to be shorter, which makes estimating the statistics of a disentangled representation difficult.
  
  In this paper, we extend VAE-DA and address the issue by learning interpretable and disentangled representations using a variant of VAEs that is designed for sequential data, named factorized hierarchical variational autoencoders (FHVAEs)~\cite{hsu2017unsupervised}.
  Instead of estimating the latent representation statistics on short utterances, we use a loss that considers the statistics across utterances in the entire corpus.
  Therefore, we can safely alter the latent part that models non-linguistic factors in order to synthesize in-domain data from out-of-domain data.
  Our proposed methods are evaluated on the AMI~\cite{carletta2007unleashing} dataset, which contains close-talking and distant-talking recordings in a conference room meeting scenario.
  We treat close-talking data as out-of-domain data and distant-talking data as in-domain data.
  In addition to outperforming all baseline methods, our proposed methods successfully close the gap between an unadapted model and a fully-supervised model by more than 77\% in terms of word error rate without the presence of any parallel data.
  
  %% SECTION 2
  \section{Limitations of Previous Work}\label{sec:issue}
  In this section, we briefly review VAE-based data augmentation and its limitations.
  
  \subsection{VAE-Based Data Augmentation}
  Generation of speech data often involves many independent factors, such as linguistic content, speaker identity, and room acoustics, that are often unobserved, or only partially observed. 
  One can describe such a generative process using a latent variable model, where a vector $\bm{z}\in\mathcal{Z}$ describing generating factors is first drawn from a prior distribution, and a speech segment $\bm{x}\in\mathcal{X}$ is then drawn from a distribution conditioned on $\bm{z}$.
  VAEs~\cite{kingma2013auto, rezende2014stochastic} are among the most successful latent variable models, which parameterize a conditional distribution, $p(\bm{x}|\bm{z})$, with a decoder neural network, and introduce an encoder neural network, $q(\bm{z}|\bm{x})$, to approximate the true posterior, $p(\bm{z}|\bm{x})$.
  
  In~\cite{hsu2017learning}, a VAE is proposed to model a generative process of speech segments.
  A latent vector in the latent space is assumed to be a linear combination of orthogonal vectors corresponding to the independent factors, such as phonetic content and speaker identity.
  In other words, we assume that $\bm{z} = \bm{z}_\ell + \bm{z}_n$ where $\bm{z}_\ell$ encodes the linguistic/phonetic content and $\bm{z}_n$ encodes the nuisance factors, and $\bm{z}_\ell \perp \bm{z}_n$.
  To augment the data set while reusing the labels, for any pair of utterance and its corresponding label sequence $(\bm{X}, y)$ in the data set, we generate $(\hat{\bm{X}}, y)$ by altering the nuisance part of $\bm{X}$ in the latent space.
  
  \subsection{Estimating Latent Nuisance Vectors}
  A key observation made in~\cite{hsu2017unsuperviseddomain} is that nuisance factors, such as speaker identity and room acoustics, are generally constant over segments within an utterance, while linguistic content changes from segment to segment.
  In other words, latent nuisance vectors $\bm{z}_n$ are relatively consistent within an utterance, while the distribution of $\bm{z}_\ell$ conditioned on an utterance can be assumed to have the same distribution as the prior.
  Therefore, suppose the prior is a diagonal Gaussian with zero mean. Given an utterance $\bm{X} = \{\bm{x}^{(n)}\}_{n=1}^N$ of $N$ segments, we have:
  \begin{align}
  	\dfrac{1}{N}\sum_{n=1}^N \bm{z}^{(n)} &= \dfrac{1}{N}\sum_{n=1}^N \bm{z}_n^{(n)} + \dfrac{1}{N}\sum_{n=1}^N \bm{z}_\ell^{(n)} \\
    &\approx \dfrac{1}{N}\sum_{n=1}^N \bm{z}_n + \mathbb{E}_{p(z)}[\bm{z}_\ell] = \bm{z}_n + 0.
  \end{align}
  That is to say, the latent nuisance vector would stand out, and the rest would cancel out, when we take the average of latent vectors over segments within an utterance.
  
\begin{figure}[t]
\centerline{\includegraphics[width=\linewidth]{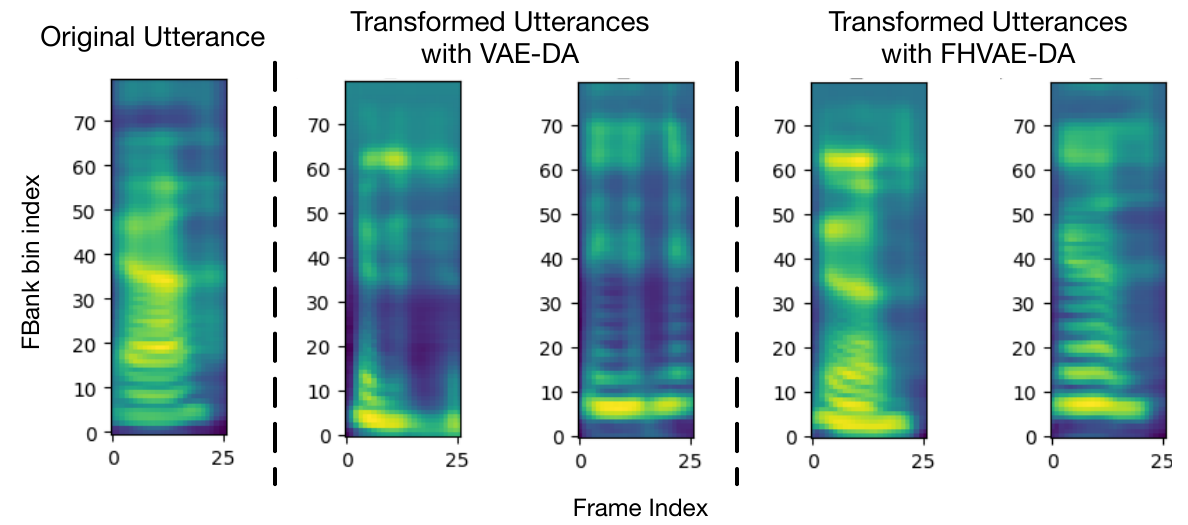}}
\caption{Comparison between VAE-DA and proposed FHVAE-DA applied to a short utterance using nuisance factor replacement. The same source and two target utterances are used for both methods. Our proposed FHVAE-DA can successfully transform only nuisance factors, while VAE-DA cannot.}
\label{fig:vae_fhvae_da}
\end{figure}

  This approach shows great success in transforming clean read speech into noisy read speech.
  However, in a conversational scenario, the portion of short utterances are much larger than that in a reading scenario.
  For instance, in the Wall Street Journal corpus~\cite{garofalo2007csr}, a read speech corpus, the average duration on the training set is 7.6s ($\pm$2.9s), with no utterance shorter than 1s.
  On the other hand, in the AMI corpus~\cite{carletta2007unleashing}, the distant conversational speech meeting corpus, the average duration on the training set is 2.6s ($\pm$2.7s), with over 35\% of the utterances being shorter than 1s.
  The small number of segments in a conversational scenario can lead to unreliable estimation of latent nuisance vectors, because the sampled mean of latent linguistic vectors would exhibit large variance from the population mean.
  The estimation under such a condition can contain information about not only nuisance factors, but also linguistic factors. 
  Indeed, we illustrate in Figure~\ref{fig:vae_fhvae_da} that modifying the estimated latent nuisance vector of a short utterance can result in undesirable changes to its linguistic content.
  
  %% SECTION 3
  \section{Methods}\label{sec:method}
  In this section, we describe the formulation of FHVAEs and explain how it can overcome the limitations of vanilla VAEs.
  
  \subsection{Learning Interpretable Disentangled Representations}
  To avoid estimating nuisance vectors on short segments, one can leverage the statistics at the corpus level, instead of at the utterance level, to disentangle generating factors.
  An FHVAE~\cite{hsu2017unsupervised} is a variant of VAEs that models a generative process of sequential data with a hierarchical graphical model.
  Specifically, an FHVAE imposes sequence-independent priors and sequence-dependent priors to two sets of latent variables, $\bm{z}_1$ and $\bm{z}_2$, respectively.
  We now formulate the process of generating a sequence $\bm{X} = \{ \bm{x}^{(n)} \}_{n=1}^N$ composed of $N$ sub-sequences:
  \begin{enumerate}
    \item an \textit{s-vector} $\bmu_2$ is drawn from $p(\bmu_2) = \mathcal{N}(\bmu_2 | \bm{0}, \sigma^2_{\bmu_2}\bm{I})$.
    \item $N$ i.i.d.\ \textit{latent segment variables} $\bm{Z}_1 = \{ \bm{z}_1^{(n)} \}_{n=1}^N$ are drawn from a global prior $p(\bm{z}_1) = \mathcal{N}(\bm{z}_1 | \bm{0}, \sigma^2_{\bm{z}_1}\bm{I})$.
    \item $N$ i.i.d.\ \textit{latent sequence variables} $\bm{Z}_2 = \{ \bm{z}_2^{(n)} \}_{n=1}^N$  are drawn from a sequence-dependent prior $p(\bm{z}_2 | \bmu_2) = \mathcal{N}(\bm{z}_2 | \bmu_2, \sigma^2_{\bm{z}_2}\bm{I})$.
    \item $N$ i.i.d.\ sub-sequences $\bm{X} = \{ \bm{x}^{(n)} \}_{n=1}^N$ are drawn from $p(\bm{x} | \bm{z}_1, \bm{z}_2) = \mathcal{N}(\bm{x} | f_{\mu_x}(\bm{z}_1, \bm{z}_2), diag(f_{\sigma^2_x}(\bm{z}_1, \bm{z}_2)))$, where $f_{\mu_x}(\cdot,\cdot)$ and $f_{\sigma^2_x}(\cdot,\cdot)$ are parameterized by a decoder neural network.
  \end{enumerate}  
  The joint probability for a sequence is formulated as follows:
  \begin{equation}
  	p(\bmu_2)\prod_{n=1}^N p(\bm{x}^{(n)} | \bm{z}_1^{(n)}, \bm{z}_2^{(n)}) p(\bm{z}_1^{(n)}) p(\bm{z}_2^{(n)} | \bmu_2). 
  \end{equation}
  
  With such a formulation, $\bm{z}_2$ is encouraged to capture generating factors that are relatively consistent within a sequence, and $\bm{z}_1$ will then capture the residual generating factors.
  Therefore, when we apply an FHVAE to model speech sequence generation, it is clear that $\bm{z}_2$ will capture the nuisance generating factors that are in general consistent within an utterance.
  
  Since the exact posterior inference is intractable, FHVAEs introduce an inference model $q(\bm{Z}_1, \bm{Z}_2, \bmu_2 | \bm{X})$ to approximate the true posterior, which is factorized as follows:
  \begin{equation}
    q(\bmu_2) \prod_{n=1}^N q(\bm{z}_1^{(n)} | \bm{x}^{(n)}, \bm{z}_2^{(n)}) q(\bm{z}_2^{(n)} | \bm{x}^{(n)}),
  \end{equation}
  where $q(\bmu_2)$, $q(\bm{z}_1 | \bm{x}, \bm{z}_2)$, and $q(\bm{z}_2 | \bm{x})$ are all diagonal Gaussian distributions.
  Two encoder networks are introduced in FHVAEs to parameterize mean and variance values of $q(\bm{z}_1 | \bm{x}, \bm{z}_2)$ and $q(\bm{z}_2 | \bm{x})$ respectively. 
  As for $q(\bmu_2)$, for testing utterances we parameterize its mean with an approximated maximum a posterior (MAP) estimation $\sum_{n=1}^N \hat{\bm{z}}_2^{(n)} / (N + \sigma^2_{\bm{z}_2} / \sigma^2_{\bmu_2})$, where $\hat{\bm{z}}_2^{(n)}$ is the inferred posterior mean of $q(\bm{z}_2^{(n)} | \bm{x}^{(n)})$;
  during training, we initialize a lookup table of posterior mean of $\bmu_2$ for each training utterance with the approximated MAP estimation, and treat the lookup table as trainable parameters. This can avoid  computing the MAP estimation of each segment for each mini-batch, and utilize the discriminative loss proposed in~\cite{hsu2017unsupervised} to encourage disentanglement.
  
  \subsection{FHVAE-Based Data Augmentation}
  With a trained FHVAE, we are able to infer disentangled latent representations that capture linguistic factors $\bm{z}_1$ and nuisance factors $\bm{z}_2$.
  To transform nuisance factors of an utterance $\bm{X}$ without changing the corresponding transcript, one only needs to perturb $\bm{Z}_2$.
  Furthermore, since each $\bm{z}_2$ within an utterance is generated conditioned on a Gaussian whose mean is $\bmu_2$, we can regard $\bmu_2$ as the representation of nuisance factors of an utterance.
  We now derive two data augmentation methods similar to those proposed in~\cite{hsu2017unsuperviseddomain}, named \textit{nuisance factor replacement} and \textit{nuisance factor perturbation}.
  
  \subsubsection{Nuisance Factor Replacement}
  Given a labeled out-of-domain utterance $(\bm{X}_{out}, y_{out})$ and an unlabeled in-domain utterance $\bm{X}_{in}$, we want to transform $\bm{X}_{out}$ to $\hat{\bm{X}}_{out}$ such that it exhibits the same nuisance factors as $\bm{X}_{in}$, while maintaining the original linguistic content.
  We can then add the synthesized labeled in-domain data $(\hat{\bm{X}}_{out}, y_{out})$ to the ASR training set.
  From the generative modeling perspective, this implies that $\bm{z}_2$ of $\bm{X}_{in}$ and $\hat{\bm{X}}_{out}$ are drawn from the same distribution.
  We carry out the same modification for the latent sequence variable of each segment of $\bm{X}_{out}$ as follows: $\hat{\bm{z}}_{2, out} = \bm{z}_{2, out} - \bm{\mu}_{2, out} + \bm{\mu}_{2, in}$, 
  where $\bmu_{2, out}$ and $\bmu_{2, in}$ are the approximate MAP estimations of $\bmu_2$.
  
  \subsubsection{Nuisance Factor Perturbation}
  Alternatively, we are also interested in synthesizing an utterance conditioned on unseen nuisance factors, for example, the interpolation of nuisance factors between two utterances. 
  We propose to draw a random perturbation vector $\bm{p}$ and compute $\hat{\bm{z}}_{2, out} = \bm{z}_{2, out} + \bm{p}$ for each segment in an utterance, in order to synthesize an utterance with perturbed nuisance factors.
  Naively, we may want to sample $\bm{p}$ from a centered isotropic Gaussian.
  However, in practice, VAE-type of models suffer from an over-pruning issue~\cite{yeung2017tackling} in that some latent variables become inactive, which we do not want to perturb.
  Instead, we only want to perturb the linear subspace which models the variation of nuisance factors between utterances.
  Therefore, we adopt a similar soft perturbation scheme as in~\cite{hsu2017unsuperviseddomain}.
  First, $\{ \bmu_2 \}_{i=1}^M$ for all $M$ utterances are estimated with the approximated MAP.  Principle component analysis is performed to obtain $D$ pairs of eigenvalue $\sigma_d$ and eigenvectors $\bm{e}_d$, where $D$ is the dimension of $\bmu_2$.
  Lastly, one random perturbation vector $\bm{p}$ is drawn for each utterance to perturb as follows:
  \begin{equation}
    \bm{p} = \gamma \sum_{d=1}^D \psi_d \sigma_d \bm{e}_d, \quad \psi_d \sim \mathcal{N}(0, 1), \label{eq:pert}
  \end{equation}
  where $\gamma$ is used to control the perturbation scale.
  
  %% SECTION 4
  \section{Experimental Setup}\label{sec:exp}
  We evaluate our proposed method on the AMI meeting corpus~\cite{carletta2007unleashing}. 
  The AMI corpus consists of 100 hours of meeting recordings in English, recorded in three different meeting rooms with different acoustic properties, and with three to five participants for each meeting that are mostly non-native speakers.
  Multiple microphones are used for recording, including individual headset microphones (IHM), and far-field microphone arrays.
  In this paper, we regard IHM recordings as out-of-domain data, whose transcripts are available, and single distant microphone (SDM) recordings as in-domain data, whose transcripts are not available, but on which we will evaluate our model.
  The recommended partition of the corpus is used, which contains an 80 hours training set, and 9 hours for a development and a test set respectively.
  FHVAE and VAE models are trained using both IHM and SDM training sets, which do not require transcripts.
ASR acoustic models are trained using augmented data and transcripts based on only the IHM training set.
  The performance of all ASR systems are evaluated on the SDM development set. The NIST asclite tool~\cite{fiscus2006multiple} is used for scoring.
  
  \subsection{VAE and FHVAE Configurations}
  Speech segments of 20 frames, represented with 80 dimensional log Mel filterbank coefficients (FBank) are used as inputs.
  We configure VAE and FHVAE models such that they have comparable modeling capacity. 
  The VAE latent variable dimension is 64, whereas the dimensions of $\bm{z}_1$ and $\bm{z}_2$ in FHVAEs are both 32.
  Both models have a two-layer LSTM decoder with 256 memory cells that predicts one frame of $\bm{x}$ at a time.
Since a FHVAE model has two encoders, while a VAE model only has one, we use a two-layer LSTM encoder with 256 memory cells for the former, and with 512 memory cells for the latter.
  All the LSTM encoders take one frame as input at each step, and the output from the last step is passed to an affine transformation
  layer that predicts the mean and the log variance of latent variables.
  The VAE model is trained to maximize the variational lower bound, and the FHVAE model is trained to maximize the discriminative segment variational lower bound proposed in~\cite{hsu2017unsupervised} with a discriminative weight $\alpha = 10$.
  In addition, the original FHVAE training~\cite{hsu2017unsupervised} is not scalable to hundreds of thousands of utterances; we therefore use the hierarchical sampling-based training algorithm proposed in~\cite{hsu2018scalable} with batches of 5,000 utterances.
  Adam~\cite{kingma2014adam} with $\beta_1 = 0.95$ and $\beta_2 = 0.999$ is used to optimize all models.
  Tensorflow~\cite{abadi2016tensorflow} is used for implementation.
  
  \subsection{ASR Configuration}
  Kaldi~\cite{povey2011kaldi} is used for feature extraction, forced alignment, decoding, and training of initial HMM-GMM models on the IHM training set. The Microsoft Cognitive Toolkit~\cite{yu2014introduction} is used for neural network acoustic model training.
  For all experiments, the same 3-layer LSTM acoustic model~\cite{sak2014long} with the architecture proposed in~\cite{zhang2016highway} is adopted, which has 1024 memory cells and a 512-node linear projection layer for each LSTM layer.
  Following the setup in~\cite{hsu2016exploiting}, LSTM acoustic models are trained with cross entropy loss, truncated back-propagation through time~\cite{williams1990efficient}, and mini-batches of 40 parallel utterances and 20 frames.
  A momentum of 0.9 is used starting from the second epoch~\cite{zhang2016highway}.
  Ten percent of training data is held out for validation, and the learning rate is halved if no improvement is observed on the validation set after an epoch.
  
  \section{Results and Discussion}\label{sec:res}
  \begin{table}
  \centering
  \caption{Baseline WERs for the AMI IHM/SDM task.}
  \resizebox{\linewidth}{!}{
  \begin{tabular}{|l|cc|}
    \hline
    & \multicolumn{2}{|c|}{WER (\%)} \\
    ASR Training Set	& SDM-dev 	& IHM-dev	\\
    \hline
    IHM			& 70.8	& 27.0 	\\
    SDM 		& 46.8 (-24.0)	& 42.5 (+15.5)	\\
    \hline
    IHM, FHVAE-DI, ($\bm{z}_1$)~\cite{hsu2018extracting} 		& 64.8 (-6.0)	& 29.0 (+2.0)	\\
    \hline
    IHM, VAE-DA, (repl)~\cite{hsu2017unsuperviseddomain}			& 62.2 (-8.0)	& 31.8 (+4.8)	\\
    IHM, VAE-DA, (p, $\gamma=1.0$)~\cite{hsu2017unsuperviseddomain}	& 61.1 (-9.7)	& 30.0 (+3.0)	\\
    IHM, VAE-DA, (p, $\gamma=1.5$)~\cite{hsu2017unsuperviseddomain}	& 61.9 (-8.9)	& 31.4 (+4.4)	\\
    \hline
  \end{tabular}
  }
  \label{tab:baseline}
  \end{table}
  
  We first establish baseline results and report the SDM (in-domain) and IHM (out-of-domain) development set word error rates (WERs) in Table~\ref{tab:baseline}.
  To avoid constantly querying the test set results, we only report WERs on the development set.
  If not otherwise mentioned, the data augmentation-based systems are evaluated on reconstructed features, and trained on a transformed IHM set, where each utterance is only transformed once, without the original copy of data.
  
  The first two rows of results show that the WER gap between the unadapted model and the model trained on in-domain data is 24\%.
  %fully supervised model is 24\%.
  The third row reports the results of training with domain invariant feature, $\bm{z}_1$, extracted with a FHVAE as is done in~\cite{hsu2018extracting}.
  It improves over the baseline by 6\% absolute.
  VAE-DA~\cite{hsu2017unsuperviseddomain} results with nuisance factor replacement (repl) and latent nuisance perturbation (p) are shown in the last three rows.
  
  \begin{table}
  \centering
  \caption{WERs of the proposed and the alternative methods.}
  \resizebox{\linewidth}{!}{
  \begin{tabular}{|l|cc|}
    \hline
    & \multicolumn{2}{|c|}{WER (\%)} \\
    ASR Training Set	& SDM-dev 	& IHM-dev	\\
    \hline
    IHM			& 70.8	& 27.0 	\\
    \hline
    IHM, FHVAE-DA, (repl)				& 59.0 (-11.8)	& 31.3 (+4.3)	\\
    IHM, FHVAE-DA, (p, $\gamma=1.0$)	& \textbf{58.6 (-12.2)}	& 30.1 (+3.1)	\\
    IHM, FHVAE-DA, (p, $\gamma=1.5$)	& 58.7 (-12.1)	& 31.4 (+4.4)	\\
    \hline
    IHM, FHVAE-DA, (rev-p, $\gamma=1.0$)	& 70.9 (+0.1) 	& 30.2 (+3.2)	\\
    IHM, FHVAE-DA, (uni-p, $\gamma=1.0$)	& 66.6 (-4.2)	& 30.9 (+3.9)	\\
    \hline
  \end{tabular}
  }
  \label{tab:fhvae_da}
  \end{table}
  
 We then examine the effectiveness of our proposed method and show the results in the second, third, and fourth rows in Table~\ref{tab:fhvae_da}.
  We observe about 12\% WER reduction on the in-domain development set for both nuisance factor perturbation (p) and nuisance factor replacement (repl), with little degradation on the out-of-domain development set.
  Both augmentation methods outperform their VAE counterparts and the domain invariant feature baseline using the same FHVAE model.
  We attribute the improvement to the better quality of the transformed IHM data, which covers the nuisance factors of the SDM data, without altering the original linguistic content.
  
  To verify the superiority of the proposed method of drawing random perturbation vectors, we compare two alternative sampling methods: \textit{rev-p} and \textit{uni-p}, similar to~\cite{hsu2017unsuperviseddomain}, with the same expected squared Euclidean norm as the proposed method. 
  The \textit{rev-p} replaces $\sigma_d$ in Eq.~\ref{eq:pert} with $\sigma_{D-d}$, where $[ \sigma_1, \cdots, \sigma_D]$ is sorted,
  while the \textit{uni-p} replaces it with $\sqrt{\sum_{d=1}^D \sigma_d^2 / D}$.
  Results shown in the last two rows in Table~\ref{tab:fhvae_da} confirm that the proposed sampling method is more effective under the same perturbation scale $\gamma = 1.0$ compared to the alternative methods as expected.
  
  \begin{table}
  \centering
  \caption{WERs on reconstructed data and original data.}
  \resizebox{\linewidth}{!}{
  \begin{tabular}{|l|cc|cc|}
    \hline
    & \multicolumn{2}{|c|}{SDM-dev WER (\%)} & \multicolumn{2}{|c|}{IHM-dev WER (\%)} \\
    ASR Training Set	& recon. & ori. 	& recon.	& ori.	\\
    \hline
    reconstruction				& 73.8	& 79.5	& 30.1	& 32.1 	\\
    \hline
    repl							& \textbf{59.0} & 71.4 & 31.3 & 34.4	\\
    \hspace{.3cm}\textit{+ori. IHM}	& 59.4 & \textbf{61.4} & \textbf{30.5} & \textbf{26.2} \\
    \hline
    p, $\gamma=1.0$				& 58.6 	& 71.8 	& 30.1 & 31.5 \\
    \hspace{.3cm}\textit{+ori. IHM}	& \textbf{58.0} & \textbf{66.2} & \textbf{29.0} & \textbf{25.9} \\
    \hline
  \end{tabular}
  }
  \label{tab:aug_ori}
  \end{table}
  Due to imperfect reconstruction using FHVAE models, some linguistic information may be lost in this process. 
  Furthermore, since VAE models tend to have overly-smoothed outputs, one can easily tell an original utterance from a reconstructed one. 
  In other words, there is another layer of domain mismatch between original data and reconstructed data.
  In Table~\ref{tab:aug_ori}, we investigate the performance of models trained with different data on both original data and reconstructed data.
  The first row, a model trained on the reconstructed IHM data serves as the baseline, from which we observe a 3.0\%/3.1\% WER increase on SDM/IHM when tested on the reconstructed data, and a further 5.7\%/2.0\% WER increase when tested on the original data.
  
  Compared to the reconstruction baseline, the proposed perturbation and replacement method both show about 15\% improvement on the reconstructed SDM data, and 8\% on the original SDM data.
  Results on the reconstructed or original IHM data are comparable to the baseline.
  The performance difference between the original and reconstructed SDM shows that FHVAEs are able to transform the IHM acoustic features closer to the reconstructed SDM data.
  We then explore adding the original IHM training data to the two transformed sets (\textit{+ori. IHM}).
  This significantly improves the performance on the original data for both SDM and IHM data sets.
  We even see an improvement from 27.0\% to 25.9\% on the IHM development set compared to the model trained on original IHM data.
  
  \begin{table}
  \centering
  \caption{Models trained on disjoint partition of IHM/SDM data.}
  \resizebox{\linewidth}{!}{
  \begin{tabular}{|l|cc|}
    \hline
    & \multicolumn{2}{|c|}{WER (\%)} \\
    ASR Training Set	& SDM-dev 	& IHM-dev	\\
    \hline
    IHM-a			& 86.5			& 31.8 	\\
    SDM-b 		& 55.4 (-31.1)	& 51.0 (+19.2)	\\
    \hline
    IHM-a, FHVAE-DA, (pert, $\gamma=1.0$)	& 62.4 (-24.1)	& 33.4 (+1.6)	\\
    \hline
  \end{tabular}
  }
  \label{tab:disjoint}
  \end{table}
  
Finally, to demonstrate that FHVAEs are not exploiting the parallel connection between the IHM and SDM data sets, we create two disjoint sets of recordings of roughly the same size, such that IHM-a and SDM-b only contain one set of recordings each.
  Results are shown in~\ref{tab:disjoint}, where the FHVAE models is trained without any parallel utterances.
  In this setting, we observe an even more significant 24.1\% absolute WER improvement from the baseline IHM-a model, which bridges the gap by over 77\% to the fully supervised model.
  
  %% SECTION 5
  \section{Conclusions and Future Work}
  \label{sec:conclusion}
  In this paper, we marry the VAE-based data augmentation method with interpretable disentangled representations learned from FHVAE models for transforming data from one domain to another.
  The proposed method outperforms both baselines, and demonstrates the ability to reduce the gap between an unadapted model and a fully supervised model by over 77\% without the presence of any parallel data.
  For future work, we plan to investigate the unsupervised data augmentation techniques for a wider range of tasks.
  In addition, data augmentation is inherently inefficient because the training time grows linearly in the amount of data we have.
  We plan to explore model-space unsupervised adaptation to combat this limitation.
  
  \eightpt
  \bibliographystyle{IEEEtran}

  \bibliography{main.bib}

\end{document}